\documentclass{article}
\usepackage{spconf,amsmath,amsfonts,graphicx,hyperref,caption,booktabs,algorithm2e}

\title{EASL: Multi-Emotion Guided Semantic Disentanglement for Expressive Sign Language Generation}

\name{Yanchao Zhao$^{3}$, Jihao Zhu$^{4}$, Yu Liu$^{1,2\dagger}$\thanks{$\dagger$Corresponding author}, Weizhuo Chen$^{1,2}$, Yuling Yang$^{1,2}$, Kun Peng$^{1,2}$}
\address{$^1$Institute of Information Engineering, Chinese Academy of Sciences\\
$^2$School of Cyber Security, University of Chinese Academy of Sciences\\
$^3$School of Rehabilitation Science and Engineering, University of Health and Rehabilitation Sciences\\  
$^4$School of Language, Literature, Music and Visual Culture, The University of Aberdeen}

\begin{document}
%\ninept
%
\maketitle
\begin{abstract}
Large language models have revolutionized sign language generation by
  automatically transforming text into high-quality sign language videos,
  providing accessible communication for the Deaf community. However, existing
  LLM-based approaches prioritize semantic accuracy while overlooking emotional
  expressions, resulting in outputs that lack naturalness and expressiveness. We
  propose EASL (Emotion-Aware Sign Language), a multi-emotion-guided
  generation architecture for fine-grained emotional integration. We introduce emotion-semantic disentanglement modules with progressive training to separately extract semantic and affective features. During pose decoding, the emotional representations guide semantic interaction to generate sign poses with 7-class emotion confidence scores, enabling emotional expression recognition. Experimental results demonstrate that EASL achieves pose accuracy superior to all compared baselines by integrating multi-emotion information and effectively adapts to diffusion models to generate expressive sign language videos.
\end{abstract}

\section{\uppercase{Introduction}}
\label{sec:introduction}
 Sign language generation represents a critical challenge in multimodal artificial intelligence, aiming to convert written or spoken language into visual expression through gestures, postures, and facial expressions\cite{shahin2024review}. With advances in deep learning and computer vision, significant progress has been made in semantic accuracy and generation quality\cite{matsuo2022deep}. Current research approaches falls into two categories: end-to-end and pose-driven approaches. The former has evolved from CNN/GAN to Transformer/diffusion models, emphasizing semantic alignment and intelligibility; the latter employs keypoint sequences to drive virtual characters\cite{zelinka2020neural}, highlighting naturalness and editability. The prevailing paradigm integrates pose-based intermediate representations with diffusion/Transformer architectures.

\begin{figure}[t]
\centering
\includegraphics[width=\linewidth]{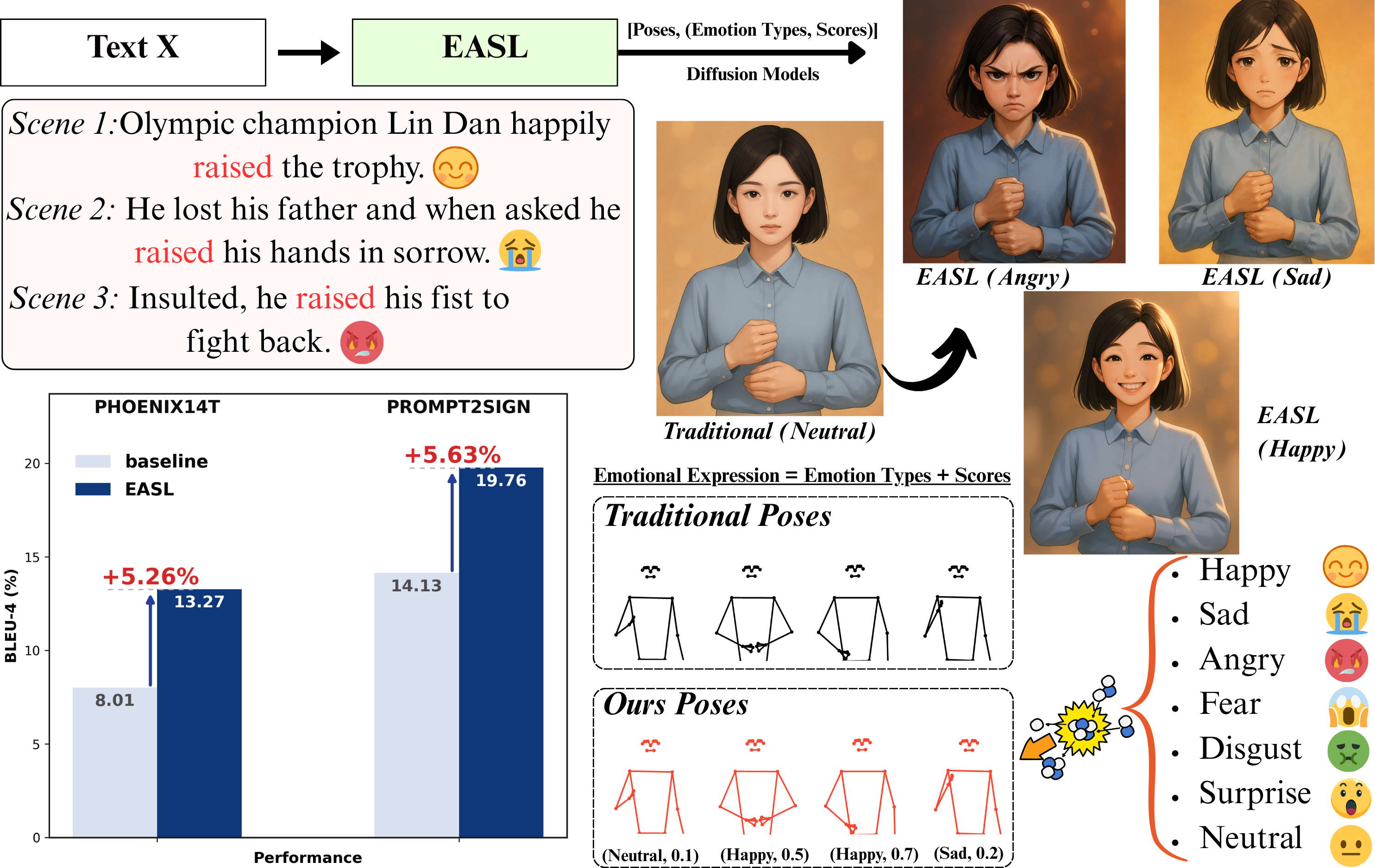}
\caption{EASL demonstrating accuracy improvements over baseline methods (\textbf{bottom left}) while applying appropriate emotional expressions to identical semantic content across different contextual scenes (\textbf{raised}) through multi-emotion guidance with confidence scores.}
\label{fig:inspire}
\end{figure}
 Despite these advances, three critical challenges persist in current sign language generation systems: \textbf{First}, insufficient emotional expression results in monotonous and unvivid generated videos that lack the natural expressiveness of human communication\cite{viegas2023including,zhang2025towards}. \textbf{Second}, inadequate integration of emotion and semantics leads to a disconnect between emotional expression and semantic content\cite{tang2022gloss,ma2024multichannel}. \textbf{Third}, reliance on static input makes it difficult to dynamically adjust emotional intensity and variability\cite{xie2024latent}, particularly in emotional scenarios where nuanced expression is crucial.

\begin{figure*}[t]
\centering
\includegraphics[width=\textwidth]{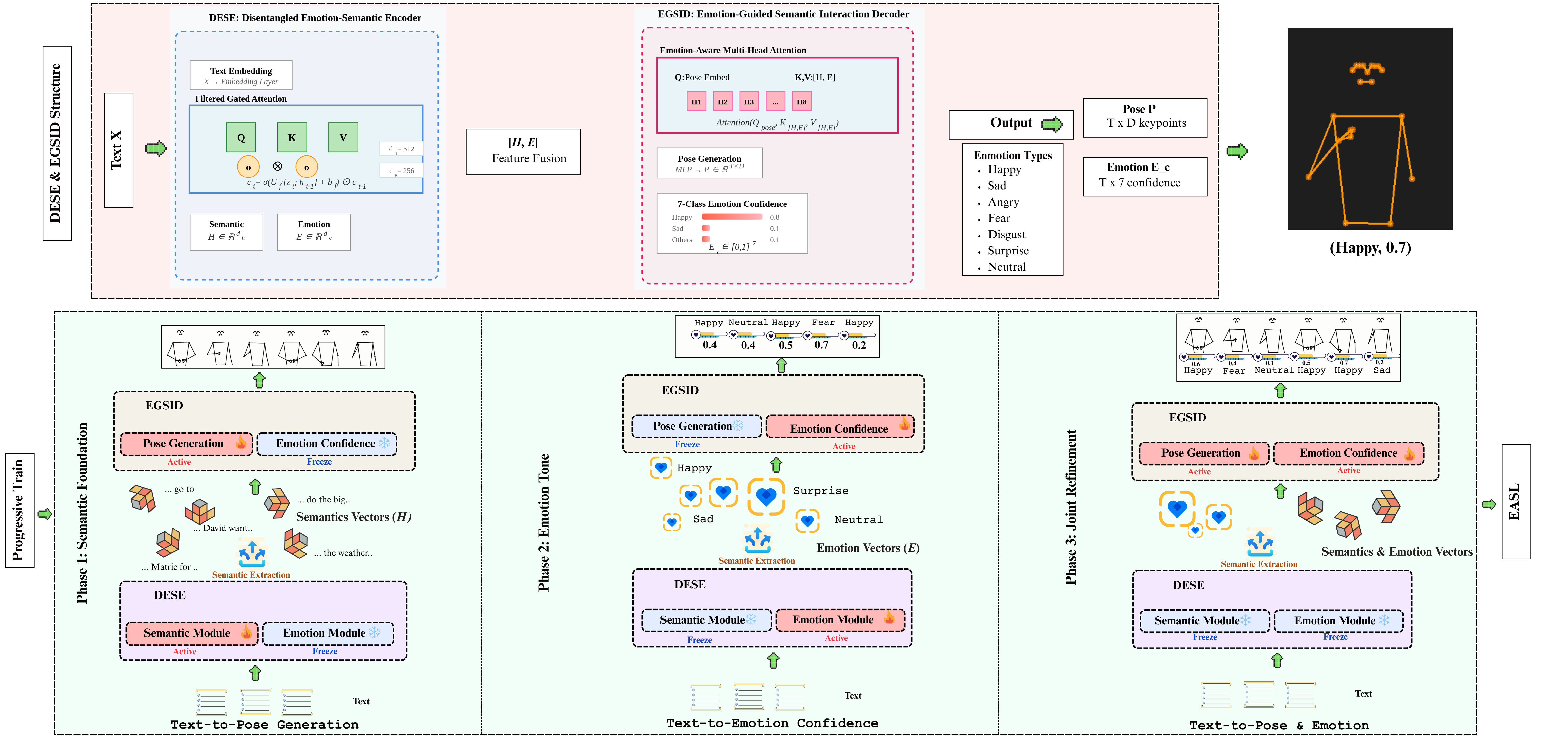}
\caption{EASL architecture with three-phase training strategy.}
\label{fig:architecture}
\end{figure*}

To address these limitations, we propose \textbf{EASL} (\textbf{E}motion-\textbf{A}ware \textbf{S}ign \textbf{L}anguage), a multi-emotion-guided, confidence-driven architecture for sign language
  generation. Our framework features two core modules: \textbf{DESE} (\textbf{D}isentangled \textbf{E}motion-\textbf{S}emantic \textbf{E}ncoder) uses gated attention to separate semantic and emotional representations, while \textbf{EGSID} (\textbf{E}motion-\textbf{G}uided \textbf{S}emantic \textbf{I}nteraction \textbf{D}ecoder) leverages emotional features to guide semantic decoding, generating poses with seven-class emotion confidence scores. A three-phase progressive training strategy sequentially builds semantic understanding, extracts emotional features, and achieves joint refinement, preventing feature entanglement while ensuring semantic accuracy and emotional expressiveness.

Our main contributions are as follows:
  \begin{enumerate}
  \item We propose \textbf{EASL\footnote{Codes: https://github.com/ycz-res/SLP/}}, a multi-emotion guided semantic disentanglement framework that decouples semantic and emotional features, effectively adapting to diffusion models to generate emotionally expressive sign language videos.
\item We design an emotion-semantic disentanglement architecture with two modules: \textbf{DESE} for separating semantic and emotional representations, and \textbf{EGSID} for emotion-guided semantic interaction.
\item We adopt three-phase progressive training strategy with parameter freezing across semantic foundation, emotional feature extraction, and joint refinement stages, achieving integration of emotional expression with sign language generation.
\item Experimental results demonstrate that EASL achieves pose accuracy superior to all compared baselines by integrating multi-emotion information and effectively adapts to diffusion models to generate expressive sign language videos.
  \end{enumerate}

\section{\uppercase{Methodology}}
\subsection{Problem Formulation}
Given text sequence $\mathcal{X} = \{x_t\}_{t=1}^T$, we aim to learn a mapping function $\mathcal{F}: \mathcal{X} \rightarrow (\mathcal{P}, \mathcal{E})$ that generates pose sequence $\mathcal{P} = \{p_i\}_{i=1}^M$ with $p_i \in \mathbb{R}^{n}$, and an emotion-confidence sequence $\mathcal{E}=\{e_i\}_{i=1}^M$, where $e_i=(e_i^{(1)},\ldots,e_i^{(K)})\in[0,1]^K$. Here, $K$ is the number of emotion categories, and the superscript $(k)$ indexes the $k$-th category.

\subsection{Architecture Overview} As shown in Figure~\ref{fig:architecture}, EASL employs a dual-module architecture with complementary streams. The upper stream maps text to pose sequences with multi-emotion confidence scores: DESE (Disentangled Emotion–Semantic Encoder) disentangles semantic and emotional representations, which EGSID (Emotion-Guided Semantic Interaction Decoder) integrates via emotion-guided semantic interactions for decoding. The lower stream illustrates the three-stage training of DESE and EGSID that underpins and optimizes the upper pipeline.

  \subsection{DESE: Disentangled Emotion-Semantic Encoder}
 DESE extracts a sequence of frame-wise semantic representations 
$H=\{h_t\}_{t=1}^{T}$ with $h_t\in\mathbb{R}^{d_h}$ (semantic dimension $d_h$) through filtered gated attention:

\begin{equation}
\tilde{c}_t = \sigma(U_f \cdot [z_t; h_{t-1}] + b_f) \odot \tilde{c}_{t-1},
\end{equation}

\begin{equation}
h_t = \mathcal{A}(W_q\tilde{c}_t,\, W_k[z_t; h_{t-1}],\, W_v[z_t; h_{t-1}]),
\end{equation}

where $\sigma$ is the sigmoid activation, $[·;·]$ denotes concatenation, and $\odot$ is element-wise multiplication. $z_t$ is the input embedding, $h_{t-1}$ the previous hidden state, and $\tilde{c}_{t-1}$ the previous filtered context. $U_f, b_f$ are gating parameters, and $W_q, W_k, W_v$ are the query/key/value projection matrices for attention $A(\cdot)$.

The emotional representation is extracted as a sequence $E = \{e_t\}_{t=1}^T$ with $e_t \in \mathbb{R}^{d_e}$ (emotion dimension $d_e$):

\begin{equation}
e_t = \sigma(W_u h_t + b_u) \odot e_{t-1},
\end{equation}

where $W_u$ and $b_u$ are the emotion-gating parameters, and $e_{t-1}$ is the previous emotional state.
Equivalently, $H\in\mathbb{R}^{T\times d_h}$ and $E\in\mathbb{R}^{T\times d_e}$.

\subsection{EGSID: Emotion-Guided Semantic Interaction Decoder}
EGSID leverages emotion-aware multi-head attention, using $E$ to guide the interaction over $H$, to generate the pose sequence $P$ and, for each frame, outputs a seven-class emotion confidence vector $E_c \in [0,1]^7$:

  \begin{equation}
  \scalebox{0.85}{$
\{P, E_c\}
= L_{\Theta}\!\Big(
A\big(U_q(P_{\mathrm{embed}} + P_{\mathrm{pos}}),\, U_k(H,E),\, U_v(H,E)\big)
\Big),$}
  \end{equation}

where $P_{\mathrm{embed}}$ and $P_{\mathrm{pos}}$ denote the query-side pose embedding and positional encoding, respectively; $U_q, U_k, U_v$ are learnable Q/K/V projections; $L_{\Theta}$ is the task-specific readout operator.

\subsection{Three-Phase Training Strategy}
To achieve emotion–semantic disentanglement and global emotion guidance, we adopt a three-stage training paradigm. Let the parameter set be $\Theta=\{\Theta_s,\Theta_e,\Theta_c\}$: $\Theta_s$includes DESE’s semantic encoder and EGSID parameters; $\Theta_e$ includes DESE’s emotion encoder and EGSID parameters; $\Theta_c$ includes only EGSID parameters. \textbf{Stage 1} (Semantic Foundation, $\Theta_s$): text→pose, learning DESE’s semantic representation. \textbf{Stage 2} (Emotion Tone, $\Theta_e$, freezing $\Theta_s$): text→multi-emotion confidence scores, learning DESE’s emotion representation. \textbf{Stage 3} (Joint Refinement, $\Theta_c$, freezing $\Theta_s$ and $\Theta_e$): train EGSID only, jointly refining text→pose and multi-emotion confidence scores.

\noindent\textbf{Loss function.} We use Mean Absolute Error (MAE) to evaluate pose and emotional generation accuracy with three main loss functions:

\begin{equation}
L_{\text{pose}} = \frac{1}{T}\sum_{t=1}^{T} \frac{1}{D}\sum_{d=1}^{D} \left| \hat{x}_{t,d} - x_{t,d} \right|,
\end{equation}

\begin{equation}
L_{\text{emo}} = \frac{1}{T}\sum_{t=1}^{T} \frac{1}{7}\sum_{k=1}^{7} \left| \hat{p}_{t,k} - p^{\text{GT}}_{t,k} \right|,
\end{equation}

\begin{equation}
L = \lambda_{\text{pose}} \cdot L_{\text{pose}} + \lambda_{\text{emo}} \cdot L_{\text{emo}},
\end{equation}

where $T$ denotes sequence length, $D$ the pose dimension, $\hat{x}$ and $x$ the predicted and ground-truth poses, $\hat{p}$ and $p^{GT}$ the predicted and ground-truth emotions, and $\lambda_{\text{pose}}$, $\lambda_{\text{emo}}$ the corresponding weights.

\begin{table*}[t]
\caption{Performance Comparison on Benchmark Datasets (BLEU1-4 and ROUGE-L).}
\centering
\footnotesize
\begin{tabular}{l|@{\hspace{0.15em}}c@{\hspace{0.15em}}c@{\hspace{0.15em}}c@{\hspace{0.15em}}c@{\hspace{0.15em}}c@{\hspace{0.15em}}}
\multicolumn{6}{c}{\textbf{(a) PHOENIX14T}} \\
\hline
Method & B-1 & B-2 & B-3 & B-4 & R-L \\
\hline
PT-base & 9.47 & 3.37 & 1.47 & 0.59 & 8.88 \\
PT-FP\&GN & 13.35 & 7.29 & 5.33 & 4.31 & 13.17 \\
DET & 17.18 & 10.39 & 7.39 & 5.76 & 17.64 \\
\underline{GEN-OBT} & \underline{23.08} & \underline{14.91} & \underline{10.48} & \underline{8.01} & \underline{23.49} \\
\textbf{EASL} & \textbf{37.46} & \textbf{20.13} & \textbf{14.97} & \textbf{13.27} & \textbf{33.00} \\
& \textbf{(+14.38)} & \textbf{(+5.22)} & \textbf{(+4.49)} & \textbf{(+5.26)} & \textbf{(+9.51)} \\
\hline
\end{tabular}
\hspace{1cm}
\begin{tabular}{l|@{\hspace{0.15em}}c@{\hspace{0.15em}}c@{\hspace{0.15em}}c@{\hspace{0.15em}}c@{\hspace{0.15em}}c@{\hspace{0.15em}}}
\multicolumn{6}{c}{\textbf{(b) Prompt2Sign}} \\
\hline
Method & B-1 & B-2 & B-3 & B-4 & R-L \\
\hline
NSLP-G & 17.55 & 11.62 & 8.21 & 5.75 & 31.98 \\
Fast-SLP & 39.46 & 23.38 & 17.35 & 12.85 & 46.89 \\
NSA & 41.31 & 25.44 & 18.25 & 13.12 & 47.55 \\
\underline{SignLLM-M} & \underline{43.40} & \underline{25.72} & \underline{19.08} & \underline{14.13} & \underline{51.57} \\
\textbf{EASL} & \textbf{50.00} & \textbf{32.61} & \textbf{24.51} & \textbf{19.76} & \textbf{47.00} \\
& \textbf{(+6.60)} & \textbf{(+6.89)} & \textbf{(+5.43)} & \textbf{(+5.63)} & \textbf{(-4.57)} \\
\hline
\end{tabular}
\label{tab:performance}
\end{table*}

\begin{table}[t]
\caption{Ablation Study Results on Test Sets.$^{\dagger}$}
\label{tab:ablation_results}
\centering
\renewcommand\arraystretch{1.2}
\small
\scalebox{0.8}{
\begin{tabular}{l|c|c|c}
\toprule
\textbf{Configuration} & \textbf{BLEU-4} & \textbf{ROUGE-L} & \textbf{MAE$^*$} \\
\midrule
\textbf{Full Model} & \textbf{16.15} & \textbf{38.69} & \textbf{5.05} \\
\midrule
w/o Three-Phases       & 12.60 (-3.55) & 33.63 (-5.06) & 8.49 (+3.44) \\
w/o $E_{DESE}$      & 14.81 (-1.34) & 35.78 (-2.91) & 6.01 (+0.96) \\
w/o $E_{{EGSID}}$  & 14.77 (-1.38) & 36.49 (-2.20) & 6.47 (+1.42) \\
w/o $E_{DESE}$ \& $E_{{EGSID}}$& 13.56 (-2.59) & 34.57 (-4.12) & 7.52 (+2.47) \\
\bottomrule
\end{tabular}
}
\footnotesize

$^{\dagger}$The current ablation experiments fully mix both datasets to simulate real-world scenarios.

$^*$MAE represents the average across 7 mixed emotion categories.

\textbf{Note:} $E_{DESE}$ denotes the emotion component in DESE; $E_{{EGSID}}$ denotes the emotion-guided multi-head attention component in EGSID.
\end{table}

\section{\uppercase{Experiments}}

\subsection{Experimental Setup}
  \noindent\textbf{Datasets.} We evaluate EASL on two benchmark sign language
    datasets: \textbf{PHOENIX14T}~\cite{koller2015continuous} and
    \textbf{Prompt2Sign}~\cite{saunders2020adversarial}. 
    
  \noindent\textbf{Baselines.} We compare EASL with four method categories: pre-trained Transformer-based methods
    (\textbf{PT-base}~\cite{saunders2020progressive},
    \textbf{PT-FP\&GN}~\cite{saunders2020progressive},
  \textbf{GEN-OBT}~\cite{tang2022gloss}), pose
    detection/regression methods (\textbf{DET}), traditional deep learning
  generation
    methods (\textbf{NSLP-G}~\cite{hwang2021non},
  \textbf{Fast-SLP}~\cite{fang2023signdiff}), and
    diffusion/generative model methods (\textbf{NSA}~\cite{baltatzis2024neural},
    \textbf{SignLLM-M}~\cite{fang2024signllm}).

  \noindent\textbf{Evaluation Metrics.} We report \textbf{BLEU-1/2/3/4} (n-gram
  precision), \textbf{ROUGE-L} (longest common subsequence), and \textbf{MAE}
  (Mean Absolute Error). BLEU-4 serves as primary
  semantic accuracy metric, while MAE evaluates emotional expression quality.
  
\noindent\textbf{Implementation Details.}
We use
  \textbf{OpenPose}~\cite{cao2017realtime} for pose
    keypoint extraction and leverage \textbf{CLIP}~\cite{radford2021learning} to
   annotate frame-level emotion
    confidence scores across seven categories (Happy, Sad, Angry, Fear, Disgust,
    Surprise, Neutral). We use two available pre-trained BERT models (semantic and emotion) to track DESE/EGSID's emotional representations and the evolution of metrics, while pose quality is evaluated via back-translation.

\subsection{Main Results}

Table~\ref{tab:performance} presents EASL's performance on datasets. Incorporating multi-emotion guidance, EASL achieves 5.26\% BLEU-4 improvement over GEN-OBT on PHOENIX14T and outperforms SignLLM-M by 5.63\% on Prompt2Sign. Lower ROUGE-L on Prompt2Sign reflects our emphasis on semantic-affective accuracy rather than word matching.

\subsection{Ablation Study}

Table~\ref{tab:ablation_results} validates each component's effectiveness. Removing the three-stage training causes the most severe degradation (BLEU-4 -3.55, ROUGE-L -5.06, MAE +3.44), confirming that progressive semantic-emotion disentanglement prevents feature entanglement. Ablating DESE's emotion branch reduces BLEU-4 by 1.34 and increases MAE by 0.96, indicating global emotion representations provide essential decoding constraints. Removing EGSID's emotion-guided attention decreases BLEU-4 by 1.38 with MAE increasing by 1.42, demonstrating its primary role in emotion regression refinement. Simultaneously removing both emotion mechanisms results in substantial performance drops (BLEU-4 -2.59, ROUGE-L -4.12, MAE +2.47), confirming their complementary necessity—DESE enables global emotional modeling while EGSID facilitates local emotional refinement, together achieving optimal semantic-emotion integration.
\subsection{Analysis}

We evaluate emotion-semantic disentanglement effectiveness using cosine similarity $\rho = \frac{1}{T}\sum_{t=1}^T 
\frac{\mathbf{z}_t^{\top}\mathbf{Z}_{\text{BERT}}}
{\|\mathbf{z}_t\|_2 \,\|\mathbf{Z}_{\text{BERT}}\|_2}$, where scores ($H-H_0$, $E-E_0$, $H-E$) are normalized to [0,1].

\begin{figure}[t]
\centering
\includegraphics[width=\linewidth]{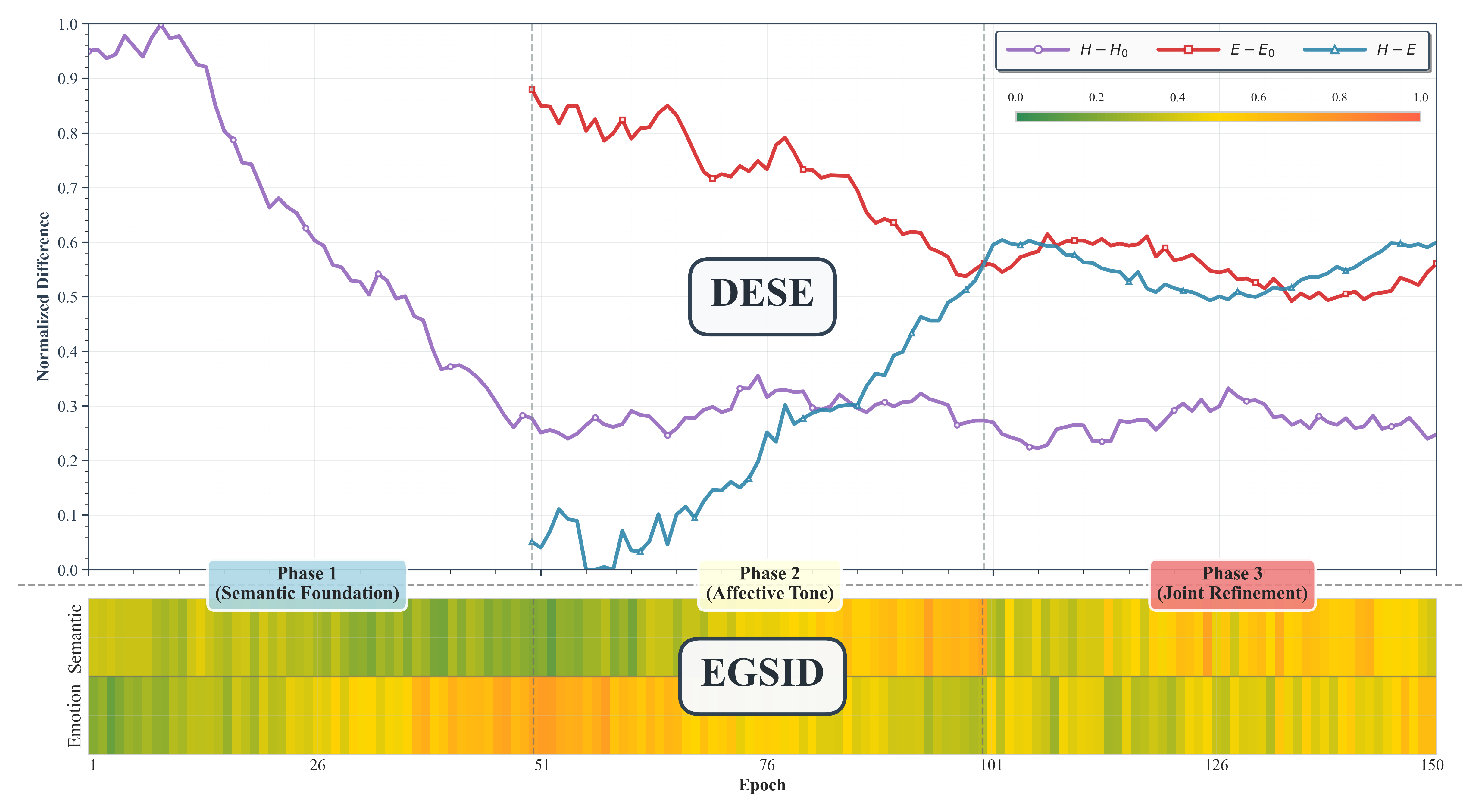}
\caption{Emotion similarity $\rho(E_t, E_{BERT})$ across epochs.}
\label{fig:similarity}
\end{figure}

Figure~\ref{fig:similarity} illustrates progressive training effectiveness. During \textbf{Phase 1}, DESE's semantic representation $H$ aligns with BERT semantic embeddings. In \textbf{Phase 2}, the emotion representation $E$ converges toward emotion-specific BERT embeddings. After freezing DESE parameters in \textbf{Phase 3}, the divergence between $H$ and $E$ stabilizes, confirming emotion-semantic disentanglement. The lower panel shows EGSID's attention patterns evolve throughout training: initially focusing on semantic features, gradually incorporating emotional signals in \textbf{Phase 2}, and achieving balanced semantic-emotion integration in the final phase. These convergence patterns validate our training strategy's effectiveness.
\subsection{Case Study}
\begin{figure}[t]
\centering
\includegraphics[width=\linewidth]{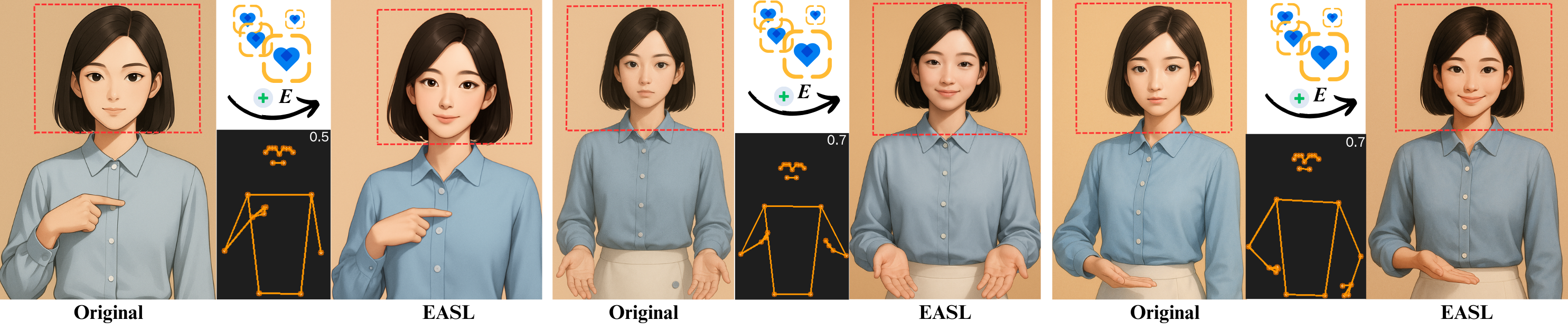}
\caption{Case: I am very pleased ... to meet you today.}
\label{cases_6}
\end{figure}

We deploy EASL to \textbf{Stable Diffusion}~\cite{rombach2022high}. Figure \ref{cases_6} shows representative frames from the generated sequences that demonstrate accurate poses and, compared to traditional methods, vividly convey the semantic context with appropriate emotional expression (Happy).

\section{\uppercase{Conclusion}}
This paper introduces EASL, a multi-emotion guided semantic disentanglement framework with progressive three-stage training for expressive sign language generation. Experimental results demonstrate EASL significantly outperforms existing methods in pose accuracy. Ablation studies validate the progressive training effectiveness and confirm the complementary roles of DESE (global emotional constraints) and EGSID (fine-grained emotion-semantic interaction). Furthermore, EASL successfully adapts to diffusion models, generating accurate poses that vividly convey semantic context with appropriate emotional expressions, significantly enhancing naturalness over traditional methods.

\newpage
\section{\uppercase{ACKNOWLEDGEMENTS}}
This research is supported by the National Key R\&D Program of China (Grant No. 2021YFF0901502). 

\bibliographystyle{IEEEbib}
\bibliography{example}

\end{document}